\pdfoutput=1

\documentclass[11pt]{article}

\usepackage[]{emnlp2021}

\usepackage{times}
\usepackage{latexsym}

\usepackage[T1]{fontenc}

\usepackage[utf8]{inputenc}

\usepackage{microtype}
\usepackage{graphicx}
\usepackage{booktabs}
\usepackage{multirow}
\usepackage{tikz}
\usepackage{amsmath}
\usepackage{array}
\usepackage{enumitem}
\usepackage{tabularx}

\def\checkmark{\tikz\fill[scale=0.4](0,.35) -- (.25,0) -- (1,.7) -- (.25,.15) -- cycle;}
\newcommand\nummicromodels{20 }

\title{Micromodels for Efficient, Explainable, and Reusable Systems: \\A Case Study on Mental Health}

\author{Andrew Lee \\
  University of Michigan \\
  \texttt{ajyl@umich.edu} \\
  \And
  Jonathan K. Kummerfeld \\ 
  University of Michigan \\
  \texttt{jkummerf@umich.edu} \\
  \AND
  Lawrence C. An \\
  University of Michigan \\
  \texttt{lcan@umich.edu} \\
  \And
  Rada Mihalcea \\
  University of Michigan \\
  \texttt{mihalcea@umich.edu}
}

\begin{document}
\maketitle

\begin{abstract}
Many statistical models have high accuracy on test benchmarks, but are not explainable, struggle in low-resource scenarios, cannot be reused for multiple tasks, and cannot easily integrate domain expertise.
These factors limit their use, particularly in settings such as mental health, where it is difficult to annotate datasets and model outputs have significant impact.
We introduce a \emph{micromodel architecture} to address these challenges.
Our approach allows researchers to build interpretable representations that embed domain knowledge and provide explanations throughout the model's decision process.
We demonstrate the idea on multiple mental health tasks: depression classification, PTSD classification, and suicidal risk assessment.
Our systems consistently produce strong results, even in low-resource scenarios, and are more interpretable than alternative methods.

\end{abstract}

\section{Introduction}
\label{sec:intro}

Systems in domains such as healthcare \citep{caruana2015intelligible} and finance
\citep{heaton2016deep}
often need to make difficult decisions that can lead to severe consequences.
Building useful systems in these settings is difficult for two key reasons: data availability and the need for explanations.
Raw data is often limited and annotating it requires specialized knowledge \citep{aguirre-etal-2021-gender}.
When a dataset is available for a task, research on models will often overfit, developing optimizations that cannot be reused for other datasets or tasks \citep{guntuku2017detecting, matero2019suicide, chen2019similar}.
Attempts to reduce data needs by integrating domain knowledge often result in inefficient and expensive models \citep{yang2019deep,liu2020k,xie2020survey}.
Integrating knowledge graphs is another alternative \citep{zhang2019ernie}, but poses challenges in domains in which domain knowledge is abstract or empirical \citep{deng2020integrating}. 
Without explanations of how these models reach their decisions, stakeholders cannot fully trust them.
In fact, despite recent advances in neural networks, it has been found that medical experts prefer simpler logistic regression models because they are more interpretable \citep{caruana2015intelligible}.

In this paper, we tackle these challenges -- explainability and reusability of models, robustness under low-resource scenarios, and integration of domain knowledge by proposing a new paradigm called a \emph{micromodel architecture}.
In this approach, a system orchestrates a collection of specialized models  
to build easily interpretable feature vectors that integrate domain knowledge.
Each micromodel is a binary classifier that represents a specific linguistic behavior.
Simple aggregators combine the output of micromodels to form a feature vector.
Finally, a task-specific model makes a prediction based on the feature vector.
Our design provides explanations along every step of its decision making process, including global and local feature importance scores, and evidence of how the input text contributes to the model's decisions.

Training this type of system involves two phases.
First, in order to build each micromodel, we introduce a data collection pipeline that uses pre-trained language models such as BERT \citep{devlin-etal-2019-bert}.
This training occurs once and then the micromodels can be reused across multiple tasks within a single domain.
Second, the task-specific model is trained on the dataset of interest.
During this phase the micromodels are not modified.

We demonstrate the benefits of micromodels in the important domain of mental health.
Recent studies have shown a rapid increase in the prevalence of depression symptoms in various demographics \citep{ettman2020prevalence}, along with elevated levels of suicidal ideation \citep{czeisler2020mental}.
Because our micromodels represent domain-level linguistic patterns, they can be reused for multiple tasks within the same domain, while requiring only half or sometimes just a quarter of the task-specific annotation data, and also having the benefit of explainability across the entire pipeline.

The primary contributions of this paper are:
(1) An efficient and reusable design using micromodels as modules to tackle various tasks within a domain by integrating domain knowledge;
(2) A data collection pipeline to build datasets for micromodels;
(3) An explainable procedure for our system's decision making process; and
(4) An analysis of the reusability and efficiency of our approach under low-resource scenarios when applied to tasks such as depression classification, PTSD classification, and suicidal risk assessment.

\section{Background and Related Work}
\label{sec:related_work}

We find inspiration in previous work that addressed explainability, reusability, efficiency under low-resource scenarios, and integration of domain expertise. We focus primarily on research that was carried out in the domain of mental health. 


\paragraph{Explainability.}
Neural networks are black-box models that lack transparency and explainability.
Structural analyses of neural networks \citep{vig2020investigating}, such as probing, has become a popular approach to investigate linguistic properties learned by language models \citep{wu2021infusing, chi-etal-2020-finding, belinkov2018evaluating, hewitt2019structural, tenney2018you}.
However, these analyses do not explain how the models use their latent information for their tasks and how they reach their decisions.
These drawbacks are especially problematic in the mental health domain \citep{carr2020ai}.
Linear models implemented with feature engineering can be analyzed via global feature importance scores, but they do not necessarily provide explanations at a query-level.
Model-agnostic explanation frameworks such as SHAP or LIME values \citep{lundberg2017unified, ribeiro2016should} can provide query-level, or local, feature importance scores, but they are approximate explanations of the underlying model.
Our approach provides (1) global and local feature importance scores, and (2) evidence from input text data that led to its output.

\paragraph{Reusability.}
Recent models in the mental health domain are often task-specific or data-specific.
Examples include features extracted from metadata \citep{guntuku2017detecting}, or neural architectures that either fine-tune their embeddings \citep{orabi2018deep} or have task-specific layers \citep{matero2019suicide}.
While task-specific designs can boost accuracy, they are difficult to extend to multiple applications.
Furthermore, \citet{harrigian2020models} show that models trained for a task in the mental health domain do not generalize across test sets that originate from different sources.
Because our micromodels are built on task-agnostic data, they are reusable for multiple applications within a domain.

\paragraph{Efficiency in Low-Resource Scenarios.}
Obtaining data in the mental health domain is difficult because of the sensitive nature of data and the need for expert annotators.
While researchers have turned to proxy-based annotations, in which data is annotated using automated mechanisms \citep{yates-etal-2017-depression, winata2018attention}, these datasets have caveats and biases \citep{aguirre-etal-2021-gender, coppersmith2015clpsych}.
These data limitations make it difficult to apply standard neural methods.

\paragraph{Integrating Domain Expertise.}
Psychologists have long studied effective methods for assessing patients for various mental health illnesses.
Assessment modules such as the Patient Health Questionnaire-9 (PHQ-9) \citep{kroenke2001phq} or PTSD Checklist (PCL) \citep{ruggiero2003psychometric} allow physicians to reliably screen for the presence or severity of various mental statuses.

Similarly, cognitive distortions are irrational or exaggerated thought patterns that can reinforce negative emotions, often exhibited by depressed patients  \citep{beck1963thinking}.
Recognizing and treating these negative thought patterns is  the focus of cognitive-behavior interventions \citep{kaplan2017cognitive}.
The PHQ-9 and an example categorization of cognitive distortions can be found in the appendix.

While these assessment modules and methods are used in clinical settings, it has been unclear how to incorporate them into automated systems.
In our work, we are able to represent responses to these questionnaires and instances of cognitive distortions using micromodels.
This allows our models to leverage domain knowledge.

\begin{figure*}[t]
\centering\includegraphics[clip, trim=0.1cm 1.2cm 0.1cm 1.0cm, width=0.95\textwidth]{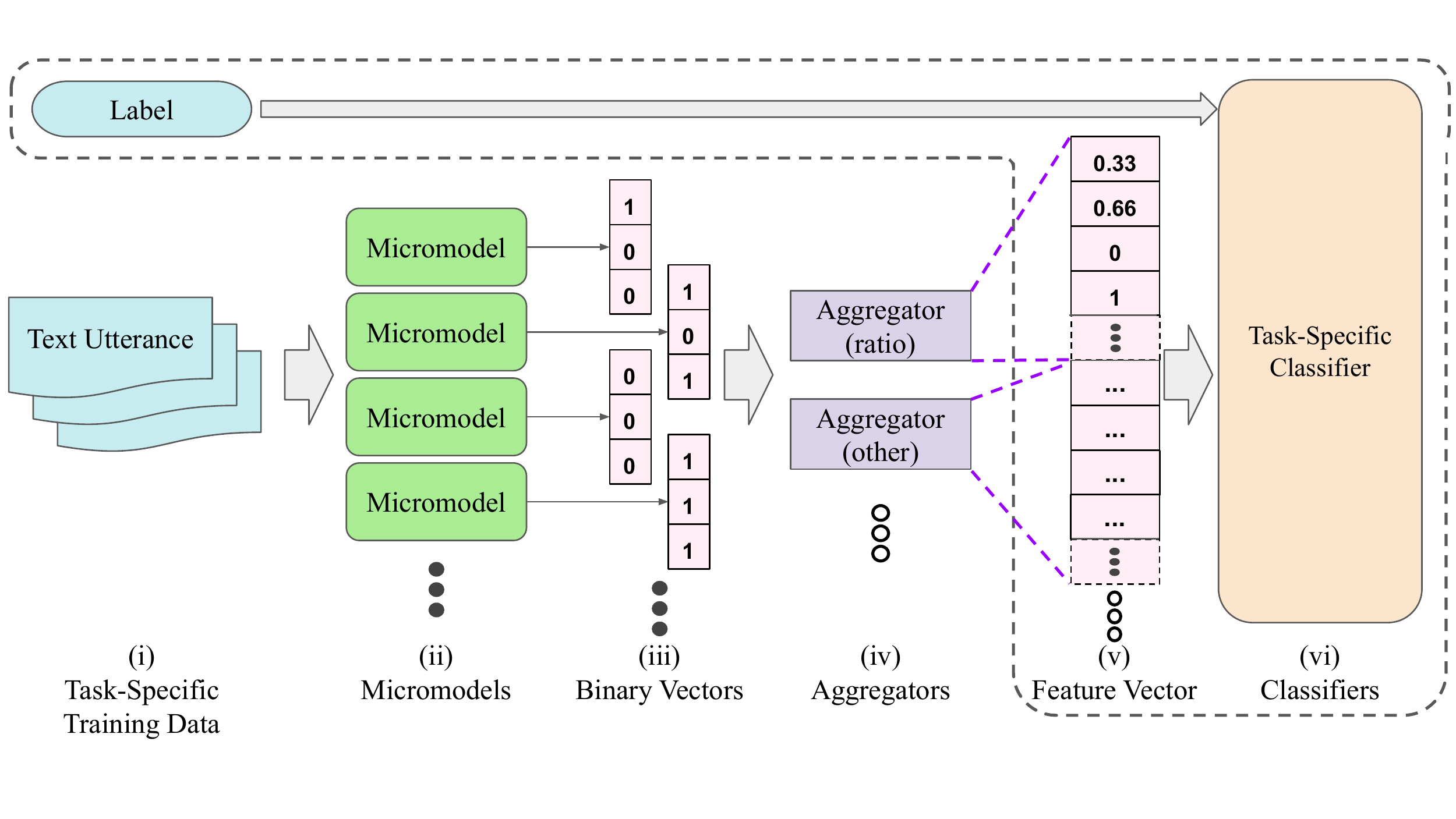}
\caption{\label{fig_system_architecture}
One training step for a task-specific classifier, given a collection of pre-built micromodels.
The input (i) is a set of utterances and a single label.
Our micromodels (ii) process each utterance to produce a set of binary vectors (iii).
Here the vectors have three elements because our example contains three utterances.
Aggregators (iv) summarize the binary vectors in a feature vector (v).
A task specific classifier takes the feature vector as input and makes a prediction, which is compared to the true label to make an update.
Note that the classifier only sees the feature vector (v) and its corresponding label for training.
}
\end{figure*}

\section{Micromodel Architecture}
\label{sec:system_architecture}

Our micromodel approach is inspired by recent work in microservice architectures---an organizational design in which applications are built from a collection of loosely coupled services \citep{nadareishvili2016microservice}.
Each of these services typically has a fine-grained focus of responsibility.
In a similar manner, we build a collection of micromodels, with  each one responsible for identifying a specific linguistic behavior.\footnote{This is where our term micromodel comes from -- each model has a fine-grained focus of responsibility. We are not referring to each model's memory footprint.}

\subsection{Micromodels}
\label{sec:sub_sec_micromodels}

A micromodel identifies a specific linguistic behavior.
We use binary classifiers for their simplicity, but our architecture is general enough to allow for other representations.
A micromodel can rely on any algorithm, from decision trees and heuristics to linear models and neural networks.

Each micromodel is responsible for representing a specific linguistic behavior.
For mental health, we developed a set of micromodels that represent examples of cognitive distortions or responses to the PHQ-9 mental health questionnaire: one micromodel identifies expressions of apathy or lack of enthusiasm (PHQ-9 question 1), while another identifies examples of all-or-nothing thinking (cognitive distortion), and so on.
We describe the process of constructing a micromodel in Section~\ref{sec:sub_sec_data_pipeline}.

\subsection{Architecture}
\label{sec:sub_sec_system_architecture}

Figure~\ref{fig_system_architecture} shows our micromodel architecture. 
At the heart of the architecture is the collection of micromodels $M = \{mm_1, ..., mm_n\}$. 
Micromodels are pre-built using a task-agnostic dataset (see Section~\ref{sec:sub_sec_data_pipeline}), and are not updated during task-specific training.\footnote{
This is an intentional choice to prevent model drift.
If we allowed updates to the task-specific models their model capacity may be repurposed to do something other than their original design intended.
}
The six steps of our architecture are:

(i) Let $(S_i, y_i)$ be one training data instance, where $S_i$ contains multiple utterances $\{s_1, ..., s_k\}$ and $y_i$ is the corresponding label for the whole set.
For instance, imagine a task of predicting a Twitter user's mental status given their recent tweets.
$S_i$ would be the user's tweets, where each $s \in S_i$ is a single tweet, and $y_i$ is the user's mental status.
Note that there are no utterance-specific labels.

(ii, iii) Given $(S_i, y_i)$, each micromodel $mm_j \in M$ produces a binary value for each utterance $s \in S_i$.
A value of 1, or a "hit", indicates that utterance $s$ is an example of the linguistic behavior that $mm_j$ is looking for.
We only use binary values in this work, but our architecture allows non-binary outputs too.
The result is $n$ binary vectors $v$ of length $k$, one from each micromodel.
Note that each binary vector $v_j$ represents the indices in $S_i$ where the target behavior of $mm_j$ can be found.

(iv, v) Each binary vector is fed through a set of aggregators.
Each aggregator maps the set of binary vectors into a feature vector that can be used for classification.
An aggregator can perform any computation.
For example, it could calculate the ratio of hits in the binary vector.
The resulting feature values would then represent the proportion of utterances that demonstrate each specific linguistic behavior.
We focus on one-to-one mappings between a micromodel and a feature value,  but they can also be many-to-one or one-to-many operations.
Together, steps (ii)-(v) could be considered a model that converts input text to a vector representation in an interpretable way.

(vi) The feature vector and its corresponding label $y_i$ are passed to a task-specific classifier.
We use explainable boosting machines (EBM) \citep{nori2019interpretml, caruana2015intelligible}, a type of generalized additive model (GAM) \citep{lou2012intelligible, lou2013accurate}.
These produce a prediction by adding together a set of functions of one or two input features.
Each function is trained using bagging and gradient boosting.
The result is a model that is more flexible than a linear model, while still being easy to interpret since it can be visualized as a set of graphs, one per function (see Section~\ref{sec:explainability} for an example of this in practice).

While the above description was used for our experiments, our framework itself is more general.

First, our micromodels are not limited to binary values (iii).
They can output continuous values, such as BERT similarity scores (Section \ref{sec:sub_sec_data_pipeline}), as long as the subsequent aggregators (iv) know how to process them.
A simple example of such aggregation might be max-pooling the micromodel output vector (iii).
In this example, the resulting feature value (v) would then represent the maximum similarity score that a micromodel identified in the task-specific training data (i).

Second, in our experiments, the task-specific classifier only sees the feature vector (v) during training, and not the original input text data.
This is not a limitation of our architecture -- other algorithms of choice could be used, including those that use neural features directly from the input text.
This may improve accuracy, but at the cost of interpretability.
Given the sensitive and high-risk domain of healthcare, where even the most accurate models become impractical without explainability \citep{caruana2015intelligible}, we use EBMs in this work.

Third, researchers can give their own definition of "Text Utterances" (i).
In the CLPsych 2015 Shard Task (Section \ref{sec:sub_sec_data}), we define each "Text Utterance" to be a single tweet from a user.
However, a different granularity could have been used, such as a set of tweets or all tweets from a user.
Such grouping allows micromodels to capture contextual information from each training data instance.

Note that only the task-specific classifier's weights are updated during training.
The micromodels are not updated -- they are only used to extract the linguistic patterns that we care about.
This is done for a few reasons:
(1) We do this to avoid each micromodel's representation from shifting away from their intended meaning;  
(2) Fine-tuning each micromodel requires labels at a micromodel granularity, rather than task-level granularity.
For instance, in the CLPsych 2015 Shared Task data (Section \ref{sec:sub_sec_data}), this means instead of (\# of users) annotations, we would need (\# of micromodels) * (\# of tweets per user) * (\# of users) annotations; and 
(3) Not all micromodels have "weights", as they can also be arbitrary heuristics (Section \ref{sec:sub_sec_micromodels}).

\begin{figure}[t]
\centering\includegraphics[clip, trim=0.25cm 0.1cm 0.7cm 0.0cm,width=0.99\columnwidth]{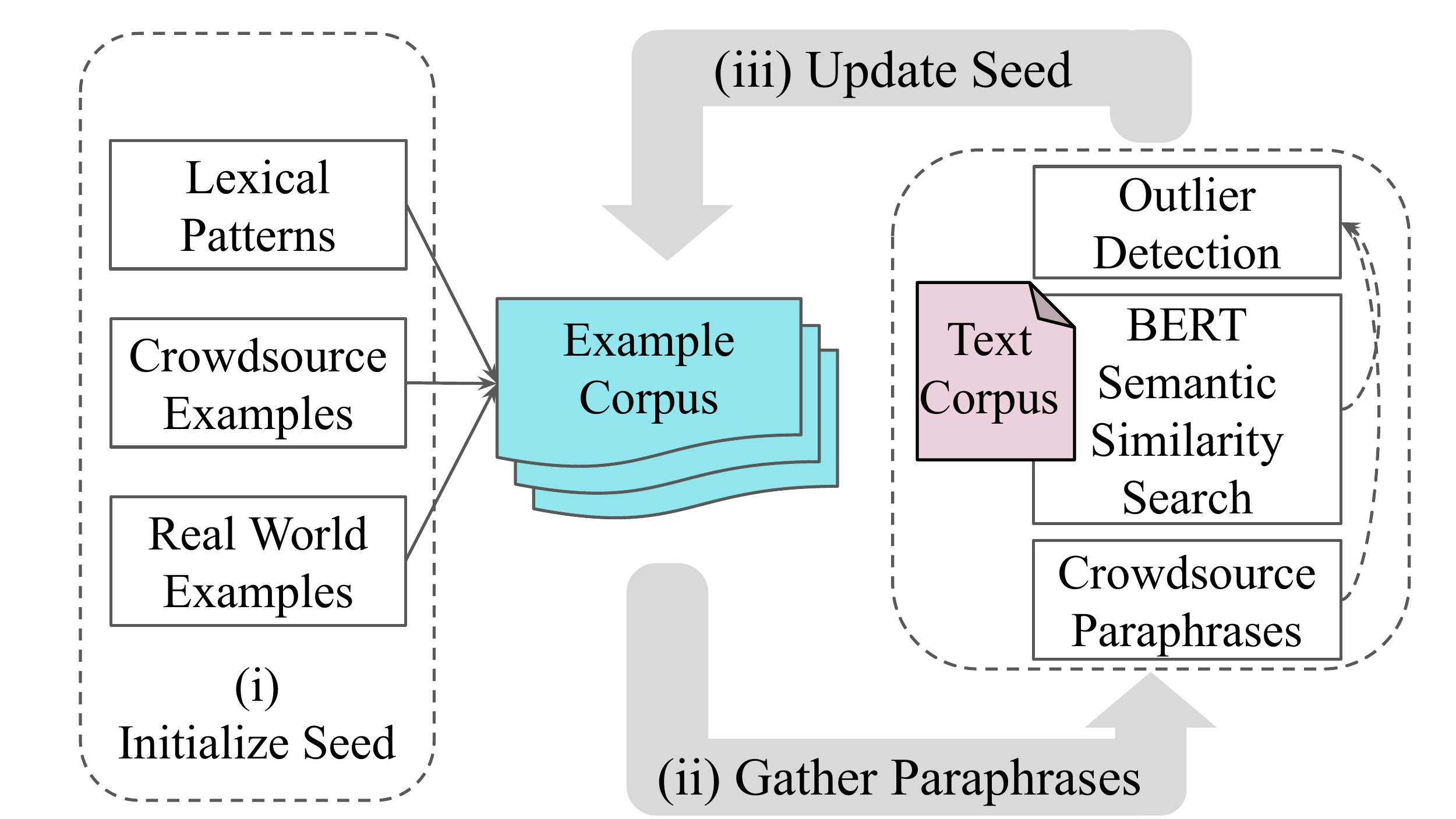}

\caption{\label{fig_data_pipeline}
Data collection pipeline for each micromodel.
Our approach is an iterative approach, in which the example corpus is updated with paraphrases.
Optionally, an outlier detection module can be incorporated in order to find the sentences that would add the most diversity to the example corpus.
}
\end{figure}

\subsection{Building Micromodels using BERT}
\label{sec:sub_sec_data_pipeline}

Each micromodel is intended to detect a specific linguistic behavior.
In order to build robust linguistic representations, it is critical to give each micromodel a diverse and representative sample of data.
However, annotating data can be time consuming and expensive.
We use BERT and Universal Sentence Encoders \citep{cer2018universal} to rapidly collect representative samples for each micromodel.
Our approach is inspired by work on collecting data for dialogue systems.
Specifically, \citet{kang2018data}, \citet{larson2019outlier}, \citet{larson-etal-2020-iterative}, and \citet{stasaski-etal-2020-diverse} proposed ways to build a diverse dataset by iteratively collecting data, starting from a seed set and crowdsourcing paraphrases.

Figure~\ref{fig_data_pipeline} depicts our pipeline for building our micromodel datasets.
For each micromodel, we build an example corpus and gather paraphrases.
While crowdsourcing can be thought of a generative approach for paraphrasing, we take a retrieval approach by using a BERT model to search for semantically similar sentences in a separate corpus of unstructured text data.
In particular, we use anonymized posts from the r/depression subreddit\footnote{\url{https://www.reddit.com/r/depression/}}, a peer support forum for anyone struggling with a depressive disorder.
While any corpus can be used to retrieve paraphrases, it is important that the linguistic phenomena that is of interest will be prevalent in the corpus.
We used Sentence Transformers \citep{reimers-gurevych-2019-sentence}\footnote{\url{https://github.com/UKPLab/sentence-transformers}} and the "paraphrase-xlm-r-multilingual-v1" pre-trained model for our semantic similarity searches.

There are multiple ways to initialize the example corpus.
One can build lexical queries by specifying patterns based on parsers or lexicons and apply them on a text corpus.
For instance, to find examples of the labeling cognitive distortion (attaching a negative label to oneself), a lexical query might look for sentences that contain a first person pronoun with a nominal subject relation with a negative token according to the LIWC lexicon \citep{pennebaker2001linguistic}.

While this may seem like an overly simple and generic pattern, because the lexical query is applied on a text corpus that pertains to depression, we are able to retrieve many examples of the target behavior, in this case the labeling cognitive distortion.
It is important to consider which text corpus the lexical query is being applied to.
To prevent micromodels from overfitting on these rule-based patterns, it is critical to run through multiple iterations of the BERT similarity search while updating the example corpus each round.
This step will identify examples of the target linguistic behavior that do not match the lexical query.

Note that this step can be pseudo-automated in a couple of ways.
One way is to apply a "negation" lexical query on the BERT results.
For instance, in the example lexical query above, given new examples of the labeling cognitive distortion according to BERT, one might apply a lexical query for utterances that do not contain a first person pronoun or a negative LIWC token.
This would identify semantically similar but syntactically diverse samples to be added back to the example corpus.

We also follow \citet{larson2019outlier} and use a Universal Sentence Encoder to identify outliers from our BERT results.
This helps us identify utterances that would add the most diversity when added back to the example corpus.
We use Snorkel\footnote{\url{https://github.com/snorkel-team/snorkel}} \citep{hancock2018training, ratner2017snorkel, ratner2016data} to construct our lexical queries.


Note that given an example corpus, applying a BERT similarity search between an input sentence and the example corpus can also be a form of a micromodel.
Once we have collected examples of a specific linguistic behavior, if the input sentence has a similarity score above a threshold value with any of the examples, our micromodel would return a value of 1, and a value of 0 otherwise. 
We call this a BERT query and use a handful of them for our experiments.
These BERT queries are able to identify examples of nuanced concepts such as cognitive distortions or a response to a PHQ-9 question, allowing us to build \emph{contextual} features that represent domain expertise.
Note that a BERT query micromodel does not require training, as we only use its inference against an example corpus.

\subsection{Discussion: Feature Engineering, Ensemble Models, and Micromodels}
\label{sec:sub_feature_eng_vs_micromodels}

Prior to neural models, many NLP systems used linear models with manually defined input features.
The process of defining these input features, sometimes called feature engineering, includes common features (e.g., unigrams, bigrams, trigrams) and domain-specific features (e.g., what time of day this tweet was posted).
One appeal of neural networks is that they can automatically learn how to combine components of the input (e.g., unigrams, timestamps) to get informative features.
While our approach has some similarities with feature engineering, there are several key differences.

First, micromodels are using \emph{external} data (such as the r/depression subreddit - Section \ref{sec:sub_sec_data_pipeline}) to learn specific linguistic phenomena.
This means they can learn things that cannot be learned from the task-specific data alone, particularly if data is limited.

Second, feature engineering typically produces a huge number of features, whereas we have on the order of tens of micromodels.
This is critical for interpretability, as we can look at the output of all our micromodels and at the patterns learned by the EBMs.
In contrast, it would be difficult to meaningfully interpret, for example, the weights assigned to all bigrams.

Third, the primary question for feature engineering is how to best summarize the available training data, while the primary questions for our approach are what data should be leveraged and what models should be built to understand and describe the training data.
Another way to view this nuance is that feature engineering extracts task-level features that suit the data for a given task.
Micromodels, on the other hand, build task-agnostic, domain-level features that can be applied on multiple tasks.

Lastly, features from prior work are typically syntactic, statistical, or derivative features, such as lexical term frequencies \citep{coppersmith2014quantifying}, extractions from metadata \citep{guntuku2017detecting}, or sentiment analyses scores \citep{chen2019similar}. 
In addition to these features, we are able to build \emph{contextual} features using contextualized language models, which are able to capture more nuanced concepts reflecting domain expertise.
While word embeddings have been used as features before \citep{mohammadi2019clac}, they are often difficult to interpret.
On the other hand, because the researcher defines the behavior of each aggregator, our resulting feature vector is easy to interpret.

Because a micromodel architecture orchestrates multiple models, it may appear similar to ensemble learning.
The key difference is that every model in an ensemble learns the same task, while the micromodels each have a different aim.
Micromodels are also intended to be used across tasks, whereas the models in an ensemble are task specific.

\section{Evaluation}
\label{sec:evaluation}

We evaluate our micromodel architecture in terms of accuracy, reusability, and efficiency under low-resource scenarios. We also address the explainability properties of our model in Section \ref{sec:explainability}. 

\subsection{Data}
\label{sec:sub_sec_data}

\paragraph{CLPsych 2015 Shared Task}~\citep{coppersmith2015clpsych}. 
This data contains tweets from 1,146 users labeled as Depression, PTSD, or Control.
Users annotated as depressed or PTSD were based on self-identified diagnosis in tweets, which were removed afterwards.
For each user identified as depressed or PTSD, an age- and gender-matched user was randomly sampled as a control user.
For each user, up to 3,000 of their most recent public tweets were collected.
The tasks include (1) classifying depression users versus control users (D vs. C), (2) classifying PTSD users versus control users (P vs. C), and (3) classifying depression users versus PTSD users (D vs. P).

\paragraph{CLPsych 2019 Shared Task}~\citep{shing2018expert, zirikly2019clpsych}.
This data is from Reddit users who have posted in the r/SuicideWatch~\footnote{\url{www.reddit.com/r/SuicideWatch/}} subreddit, a peer support forum for anyone struggling with suicidal thoughts, and were annotated with 4 levels of suicidal risk (no risk, low, moderate, severe).
A group of users who have never posted on r/SuicideWatch was used as a control group.
The shared task includes 3 tasks:
Task A is risk assessment looking \emph{only} at the users' posts in r/SuicideWatch.
Task B is also risk assessment, but also provides posts across other subreddits.
Task C is about screening, with only posts that are \emph{not} in r/SuicideWatch available, which removes self-reported evidence of risk.

\begin{table*}
\resizebox{\textwidth}{!}{
    \centering
    \begin{tabular}{lllrl}
        \toprule
        Name       & Algorithm & Category & Corpus Size & Reference that motivated this micromodel  \\
        \midrule
        All-or-Nothing Thinking     & SVM           & Cognitive Distortion  &   - & \citet{beck1963thinking, bridges2010role} \\
        Labeling                    & BERT Query    & Cognitive Distortion  & 106 & \citet{beck1963thinking, bridges2010role} \\
        Fortune-Telling Error       & BERT Query    & Cognitive Distortion  & 220 & \citet{beck1963thinking, bridges2010role, sastre2021jumping} \\
        Loss of Concentration       & BERT Query    & PHQ-9                 &  38 & \citet{kroenke2001phq} \\
        Feeling Down, Depressed     & BERT Query    & PHQ-9                 & 195 & \citet{kroenke2001phq} \\
        Poor Appetite or Overeating & BERT Query    & PHQ-9                 &  49 & \citet{kroenke2001phq}  \\
        Self Harm                   & BERT Query    & PHQ-9                 &  54 & \citet{kroenke2001phq}  \\
        Feeling Worried, Nervous, Anxious   & BERT Query    & GAD-7         &  66 & \citet{spitzer2006brief} \\
        Diagnosis                   & BERT Query    & Other                 &  55 & \\
        Self-Blaming                & BERT Query    & Other                 &  37 & \citet{zahn2015role} \\
        Substance Abuse             & BERT Query    & Other                 & 109 & \citet{abraham1999order, levy1989suicidality} \\
        Victimhood                  & BERT Query    & Other                 &  73 & \citet{swearer2001psychosocial} \\
        Mental Illness Keywords     & Logic         & Other                 &   - & \citet{cohan2018smhd} \\
        Antidepressants Keywords    & Logic         & Other                 &   - & \\
        Depression Keywords         & Logic         & Other                 &   - & \\
        PTSD Keywords               & Logic         & Other                 &   - & \\
        LIWC Sadness                & Logic         & Other                 &   - & \citet{cohan2018smhd} \\
        LIWC Anger                  & Logic         & Other                 &   - & \citet{cohan2018smhd} \\
        LIWC Joy                    & Logic         & Other                 &   - & \citet{cohan2018smhd} \\
        LIWC Fear                   & Logic         & Other                 &   - & \citet{cohan2018smhd} \\
        \bottomrule
    \end{tabular}}
    \caption{\label{tab:micromodel_detailed}
    The micromodels we developed for this work.
}
\end{table*}

\subsection{Experimental Setup}
\label{sec:sub_sec_model}

We use \nummicromodels micromodels consisting of algorithms such as SVM, BERT queries, as well as heuristics.
The choices for our micromodels were mainly motivated by existing tools commonly used by practitioners in the mental health domain, such as the PHQ-9 questionnaire and cognitive distortions.
Out of the PHQ-9 questions and cognitive distortions, those with abundant examples in the r/depression subreddit were built as micromodels.
Other linguistic behaviors that practitioners have studied \citep{zahn2015role, abraham1999order, levy1989suicidality, swearer2001psychosocial, cohan2018smhd} were included as well.
Details about each micromodel can be found in Table~\ref{tab:micromodel_detailed}.
For our SVM micromodel, we use a linear kernel and a bag of words feature representation \footnote{\url{https://scikit-learn.org/stable/modules/generated/sklearn.svm.SVC.html}}.
Our Mental Illness, Antidepressants, Depression, and PTSD keyword micromodels use a carefully curated mapping of health conditions to n-grams\footnote{\url{https://github.com/kharrigian/mental-health-keywords}}, which were extracted from \citet{benton-etal-2017-multitask}, and simply return 1 if any corresponding keywords are found in the input utterance.
Similarly, our LIWC micromodels return 1 when a keyword for each emotion is found according to LIWC.
Each BERT query micromodel has its own example corpus built using our data collection pipeline (Section~\ref{sec:sub_sec_data_pipeline}), and uses a similarity score threshold value of 0.85.
We use two aggregators.
One is as described in Section~\ref{sec:sub_sec_system_architecture}, which returns the ratio of hits in a binary vector.
The other aggregator looks for "windows": segments within each binary vector where many hits occur close to one another.
These windows may represent temporal "episodes" -- for instance, a period in which someone felt apathetic (PHQ-9 question 1), or a period in which someone had a sleeping disorder (PHQ-9 question 3) and so on.

\subsection{Results and Analyses}
\label{sec:sub_sec_results}

\begin{figure*}[t]
\centering\includegraphics[clip, trim=0.25cm 0.25cm 0.25cm 0.25cm, width=0.95\textwidth]{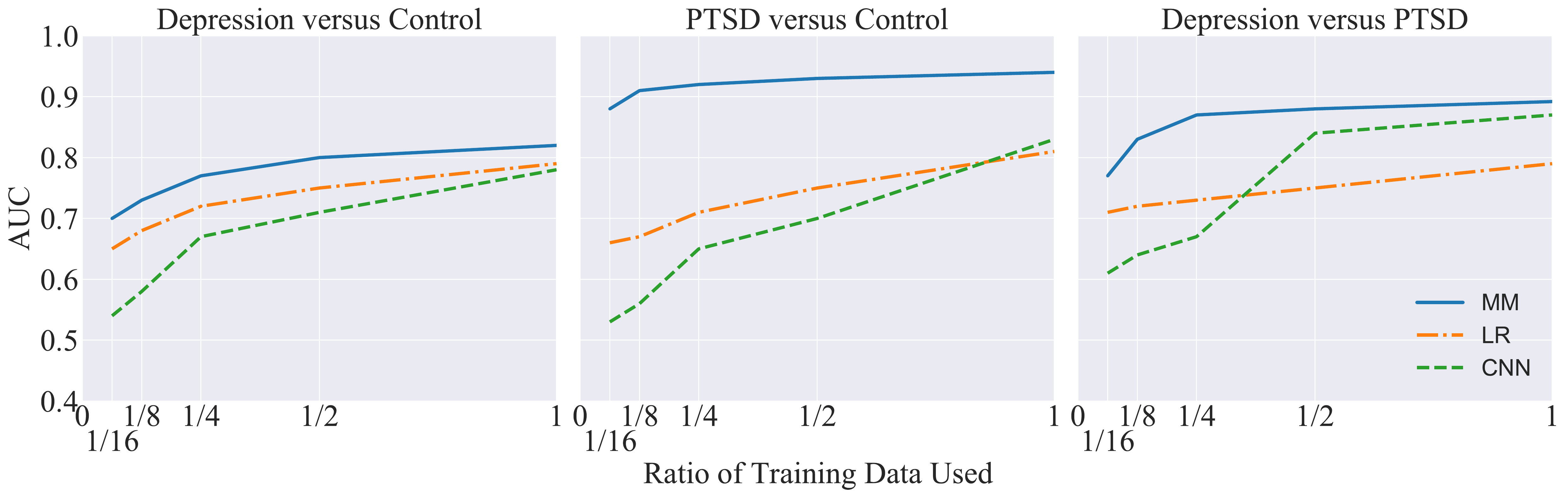}
\caption{\label{fig_low_resource}
AUC scores for various models under low-resource scenarios.
Each curve is an average of 5 runs, with random samples of training data for each run.
Our micromodel approach (MM) converges in performance with half, and sometimes just a quarter of the task-specific annotation data.
While logistic regression (LR) sees a more linear improvement and a convolutional neural network (CNN) sees sporadic jumps in performance, our approach flattens out early, indicating early convergence and less reliance on the annotated data. 
}
\end{figure*}

\begin{table}\footnotesize
\begin{tabular}{l ccccc}
    \toprule
    \multicolumn{1}{c}{Model} & \multicolumn{1}{c}{Expl?} & \multicolumn{1}{c}{Reuse?} & \multicolumn{1}{c}{D vs C} & \multicolumn{1}{c}{P vs C} & \multicolumn{1}{c}{D vs P} \\
    
     & & & \multicolumn{1}{c}{n = 654} & \multicolumn{1}{c}{n = 492} & \multicolumn{1}{c}{n=573} \\
     
    \midrule
    LR   & \checkmark & \checkmark & 0.8            & 0.817          & 0.785 \\
    CNN  &            & \checkmark & 0.79           & 0.85           & 0.87  \\
    UMD  &            &            & 0.86           & 0.893          & 0.841 \\
    WWBP &            &            & \textbf{0.904} & 0.916          & 0.81  \\
    MM   & \checkmark & \checkmark & 0.821          & \textbf{0.936} & \textbf{0.892} \\
    \bottomrule
\end{tabular}
\caption{\label{tab:clpsych2015_depression_vs_control}
AUC scores for various approaches, where LR is a logistic regression model, CNN is a convolutional neural network, and MM is our micromodel approach.
UMD is from \citet{resnik2015university}, WWBP is from \citet{preotiuc2015mental} -- these two systems were the only ones that reported AUC scores and are directly comparable to ours.
We also indicate whether each approach is explainable and reusable.
}
\end{table} 

\paragraph{Accuracy.}
We follow prior work \citep{resnik2015university, preotiuc2015mental} and use ROC area-under-the-curve (AUC) to evaluate the accuracy of our approach, along with a wide range of baseline models and present them in Table~\ref{tab:clpsych2015_depression_vs_control}.
We include a logistic regression model, which has been a simple yet effective benchmark in similar tasks \citep{harrigian2020models}, as well as a convolution neural network (CNN) based on \citet{orabi2018deep}.
Lastly we include any AUC scores that were available from system submissions from the shared task.
Our approach consistently demonstrates high AUC scores, with the highest AUC scores for classifying PTSD users against control users and depression users against PTSD users.

\paragraph{Efficiency in Low-Resource Scenarios.}
Gathering and annotating data can be both time consuming and expensive, especially within the mental health domain.
Our approach can work with relatively little task-specific data.
The micromodels are not retrained, and the task-specific classifier can work with limited data because it is (1) a relatively simple model, and (2) informed by the micromodels.
Figure~\ref{fig_low_resource} shows the AUC scores of our approach compared to our baseline models with various amounts of task-specific annotated data.
We consider five sets; the first has a random sample of 1/16th of the available training data, and each subsequent set has twice as much data.
We show results averaged over five runs of this data sampling process.
Unlike the baseline models, our approach stays robust down to just 1/4th of the training data.


\paragraph{Reusability.}
Because micromodels are task agnostic, they can be reused for tasks within the same domain.
This contrasts with the standard way of developing models, where the annotation scheme, embeddings, model structure, and so on, are carefully designed, curated, or fine-tuned per task.

In order to demonstrate the reusability of our micromodels, we also apply them to the CLPsych 2019 shared task.
Note that none of the micromodels were updated -- only the weights for the EBM classifier were learned using the annotated data.
Table~\ref{tab:clpsych2019_suicide_risk_assessment} shows the macro-$F_1$ scores of our approach amongst the systems submitted to the shared task\footnote{We exclude systems without paper submissions}.
Because we care about reusability, they are sorted by their average ranking across the three tasks.

There are a couple of observations to make from these results.
First, despite not having any task-specific design in place, our approach ranks 3rd amongst the systems on average.
Second, our approach is one of the best performing approaches for Task C.
Unlike the first two assessment tasks, Task C is concerned with screening for suicidal risk given \emph{none} of their posts from r/SuicideWatch.
Because of the lack of self-reported evidence of any suicidal ideation, this task was considered the hardest task, as evident by the low $F_1$ scores.
Since our suite of micromodels are built to identify various linguistic traits of depressive users, even without immediate signals of suicidal ideation, our approach is able to detect signs of depression, a precursor for suicide risk, and screen for users with potential risk of suicide.
We believe this demonstrates our micromodels' ability to understand domain-level concepts, rather than task-specific patterns, thus allowing our micromodels to be reused in multiple tasks within the same domain.

\begin{table} \footnotesize
\begin{tabular*}{0.98\columnwidth}{l ccc}

\toprule
                              & \multicolumn{3}{c}{r/SuicideWatch Data?} \\
    & \multicolumn{1}{c}{Only}    & \multicolumn{1}{c}{Yes}    & \multicolumn{1}{c}{No}    \\
    \multicolumn{1}{c}{Model} & (Task A) & (Task B) & (Task C) \\
    \midrule
    
    \citeauthor{mohammadi2019clac}        & \textbf{0.481}$\bullet$ & 0.339$\bullet$          & \textbf{0.268}$\bullet$ \\
    \citeauthor{matero2019suicide}        & 0.459$\bullet$          & \textbf{0.457}$\bullet$ & 0.176                 \\
    \emph{Micromodels}                    & \textcolor{blue}{0.395} & \textcolor{blue}{0.274} & \textcolor{blue}{0.255} \\
    \citeauthor{ambalavanan2019using}     & 0.477$\bullet$          & 0.261                   & 0.159                 \\
    \citeauthor{rissola2019suicide}       & 0.291                   & 0.311$\bullet$          & 0.136                 \\
    \citeauthor{morales2019investigation} & 0.178                   & 0.212                   & 0.165                 \\
    \citeauthor{iserman2019dictionaries}  & 0.402$\bullet$          & 0.148                   & 0.118                 \\
    \citeauthor{bitew2019predicting}      & 0.445$\bullet$          & -                       & -                     \\
    \citeauthor{allen2019convsent}        & 0.373                   & -                       & -                     \\
    \citeauthor{hevia2019analyzing}       & 0.312                   & -                       & -                     \\
    \citeauthor{ruiz2019clpsych2019}      & -                       & 0.370$\bullet$          & -                     \\
    \citeauthor{chen2019similar}          & -                       & 0.358$\bullet$          & -                     \\
    
    \bottomrule
\end{tabular*}
\caption{\label{tab:clpsych2019_suicide_risk_assessment}
Macro-$F_1$ scores of micromodels and system submissions from the CLPsych 2019 Shared Task.
To understand the reusability of each system across the three tasks, they are sorted by the average of their rankings on each task.
$\bullet$  indicates scores higher than that of our approach.
}
\end{table} 

\section{Step-wise Explanations}
\label{sec:explainability}

Our micromodel architecture provides various levels of explanations during each step. 
We first demonstrate the explanations provided by EBM classifiers before walking through each step. 

\begin{figure}
\centering\includegraphics[width=0.95\columnwidth]{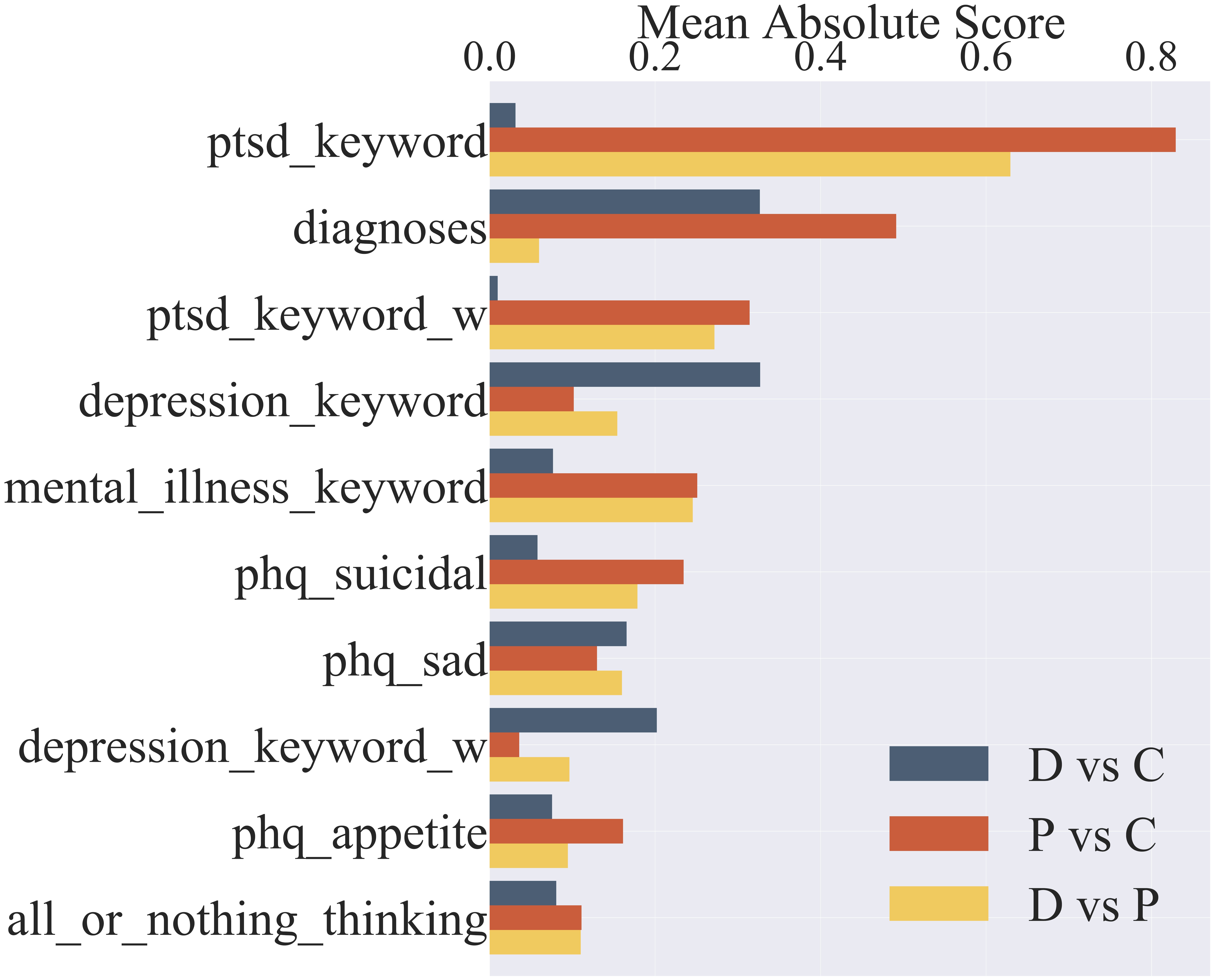}
\caption{\label{fig_feature_importance}
Ten most important features according to their average global feature importance scores on the three CLPsych 2015 shared tasks: depression versus control, PTSD versus control, and depression versus PTSD.
Features ending in "w" are features from the aggregator that looks for windows of hits.
}
\end{figure}

EBMs are additive models in which a nonlinear function $f_i$ is learned for each input feature $i$.
One can calculate global feature importance scores by applying each feature function $f_i$ on every point $t$ in the training data.
We then take the average of the absolute value of $f_i(t)$ for each feature $i$:

\setlength{\abovedisplayskip}{-5pt}
\setlength{\abovedisplayshortskip}{-5pt}
\setlength{\belowdisplayskip}{3pt}
\setlength{\belowdisplayshortskip}{3pt}
\small
\begin{equation}
Feature Importance_i = avg(abs(f_i(t))), t \in T
\end{equation}
\normalsize

\noindent where $T$ is our entire training data.
Figure~\ref{fig_feature_importance} shows the top 10 most important features for the three CLPsych 2015 shared tasks.
Similarly, we can explain the model's decision for a specific instance $t \in T$ by simply applying $f_i(t)$ for each $i$.

\begin{figure}
\centering\includegraphics[clip, trim=3.75cm 0.0cm 0.15cm 0.0cm, width=0.95\columnwidth]{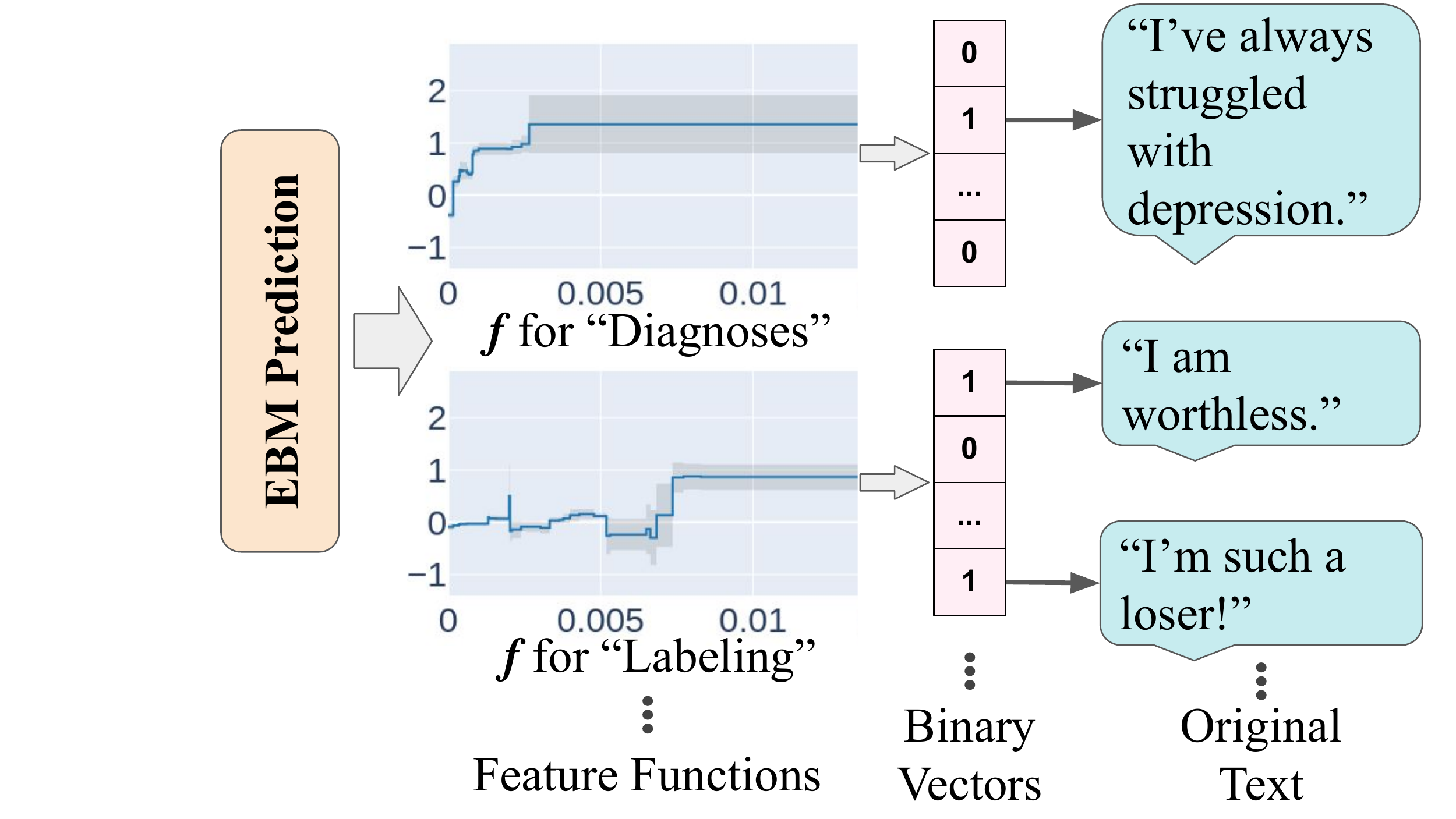}
\caption{\label{fig_explain_process}
Explanations provided by various steps in the micromodel architecture.
The feature functions $f$ provide details on how each feature contributes to the classifier's decisions.
The binary vectors indicate the location of each micromodel's hits in the input text data, allowing us to look them up as evidence. 
}
\end{figure}

Inspecting the plots of each $f_i$ also provides a granular explanation of our classifier.
Figure~\ref{fig_explain_process} contains examples of two of the feature functions from the depression detection task.
The x-axis indicates the ratio of hits for each micromodel -- in the context of this task, this represents the ratio of tweets per user that contain a specific linguistic behavior.
While $f_{\text{Diagnoses}}$ produces a strong signal when a user contains \emph{any} tweets that exhibit a diagnosis statement, $f_{\text{Labeling}}$ produces a strong signal when more than roughly 0.75\% of a user's tweets contain an example of the labeling cognitive distortion.

Other than the EBM classifier, our approach also provides explanations throughout each step.
The first step consists of the micromodels, whose explainability depends on their underlying algorithms.
The choice of these models likely involves a trade-off between accuracy and explainability.

The binary vectors produced by the micromodels indicate the utterance in which a specific linguistic behavior can be found.
This provides provenance for our feature vector -- we can use them to look up the sentences in the original input text before they were featurized.
Figure~\ref{fig_explain_process} demonstrates this process~\footnote{We use fabricated examples to protect the identity of Twitter users in the dataset.}.
Such text data provides evidence for the model's decisions.
This text data can be combined with the feature importance scores to understand how they affected the model's decisions, or to uncover patterns in the users' behaviors.

As for aggregators, in this work we use simple and intuitive operations, making the resulting feature vector easy to interpret.
Note that without an interpretable feature vector, the feature functions $f_i$ also become difficult to understand as well.

\section{Conclusion}
\label{sec:conclusion}

In this paper, we introduced a new framework that uses a collection of micromodels to tackle various tasks within the mental health domain.
Rather than directly applying contextualized language models to a task, we use them to rapidly collect diverse samples to build micromodels, which leads to a distributed-learning paradigm.
Incorporating contextual language models in our data collection allows us to capture nuanced behaviors such as cognitive distortions.
Furthermore, our pipeline allows us to leverage any amount of external data, rather than extracting features within the task domain.

The resulting micromodels allow us to build \emph{contextual} features, each of which can represent linguistic behaviors or domain knowledge.
Such a feature vector is intuitive to interpret while being effective for classifiers to learn from, even in low-resource scenarios in which not a lot of task-specific annotation data is available.
Our approach provides explanations throughout the entire decision making process, including both global and local feature importance scores, as well as the exact locations of the text that contributed to the model's decisions.
Because our micromodels are built in a task-agnostic manner, they can be reused for multiple tasks within the same domain.

The code for our micromodel architecture is publicly available at \url{https://github.com/MichiganNLP/micromodels.git}.

\section{Ethical Considerations}
\label{sec:ethics}

While we believe our approach takes a step towards the application of intelligent systems to data-poor or sensitive domains such as mental health, it is important to discuss potential risks, harm, and limitations of our work.

Because our approach heavily relies on micromodels that represent linguistic behaviors or domain knowledge, it is critical that their representations are faithful.
The authors responsible for building our micromodels were trained on cognitive behavior therapy and cognitive distortions.
It is important to have trained experts heavily involved throughout our data collection process and guiding the evaluation of how accurate the micromodels are.
This leads to a limitation of our work.
While we evaluated our approach in an end-to-end manner for various tasks, we found it challenging to evaluate the micromodels in isolation.
The difficulty in building test sets arise from not only the effort involved in gathering accurate annotations, but also from requiring high coverage and diversity of linguistic phenomena in the data as well.

Lastly, \citet{aguirre-etal-2021-gender} demonstrate that the CLPsych 2015 shared task dataset is not demographically representative.
Our work is only a proof of a concept, and to be applied in a real world scenario, a non-biased dataset should be used.

\section*{Acknowledgements}

We would like to thank Joseph Himle, Addie Weaver, and Anao Zhang from the School of Social Work at University of Michigan for the training on cognitive behavior therapy and cognitive distortions.
We thank the members of the LIT lab at University of Michigan for constructive feedback.
We thank the EMNLP reviewers for their helpful suggestions. 
This material is based in part upon work supported by a Google focus award, by the Precision Health initiative at the University of Michigan, and by DARPA (grant \#D19AP00079).

\bibliography{anthology,custom}

\begin{thebibliography}{69}
\expandafter\ifx\csname natexlab\endcsname\relax\def\natexlab#1{#1}\fi

\bibitem[{Abraham and Fava(1999)}]{abraham1999order}
Henry~David Abraham and Maurizio Fava. 1999.
\newblock \href {https://pubmed.ncbi.nlm.nih.gov/9924877/} {Order of onset of
  substance abuse and depression in a sample of depressed outpatients}.
\newblock \emph{Comprehensive Psychiatry}, 40(1):44--50.

\bibitem[{Aguirre et~al.(2021)Aguirre, Harrigian, and
  Dredze}]{aguirre-etal-2021-gender}
Carlos Aguirre, Keith Harrigian, and Mark Dredze. 2021.
\newblock \href {https://www.aclweb.org/anthology/2021.eacl-main.256} {Gender
  and racial fairness in depression research using social media}.
\newblock In \emph{Proceedings of the 16th Conference of the European Chapter
  of the Association for Computational Linguistics: Main Volume}, pages
  2932--2949.

\bibitem[{Allen et~al.(2019)Allen, Bagroy, Davis, and
  Krishnamurti}]{allen2019convsent}
Kristen Allen, Shrey Bagroy, Alex Davis, and Tamar Krishnamurti. 2019.
\newblock \href {https://www.aclweb.org/anthology/W19-3024/} {Convsent at
  clpsych 2019 task a: Using post-level sentiment features for suicide risk
  prediction on reddit}.
\newblock In \emph{Proceedings of the Sixth Workshop on Computational
  Linguistics and Clinical Psychology}, pages 182--187.

\bibitem[{Ambalavanan et~al.(2019)Ambalavanan, Jagtap, Adhya, and
  Devarakonda}]{ambalavanan2019using}
Ashwin~Karthik Ambalavanan, Pranjali~Dileep Jagtap, Soumya Adhya, and Murthy
  Devarakonda. 2019.
\newblock \href {https://www.aclweb.org/anthology/W19-3022/} {Using contextual
  representations for suicide risk assessment from internet forums}.
\newblock In \emph{Proceedings of the Sixth Workshop on Computational
  Linguistics and Clinical Psychology}, pages 172--176.

\bibitem[{Beck(1963)}]{beck1963thinking}
Aaron~T Beck. 1963.
\newblock \href
  {https://jamanetwork.com/journals/jamapsychiatry/article-abstract/488402}
  {Thinking and depression: I. idiosyncratic content and cognitive
  distortions}.
\newblock \emph{Archives of general psychiatry}, 9(4):324--333.

\bibitem[{Belinkov et~al.(2018)Belinkov, M{\`a}rquez, Sajjad, Durrani, Dalvi,
  and Glass}]{belinkov2018evaluating}
Yonatan Belinkov, Llu{\'\i}s M{\`a}rquez, Hassan Sajjad, Nadir Durrani, Fahim
  Dalvi, and James Glass. 2018.
\newblock Evaluating layers of representation in neural machine translation on
  part-of-speech and semantic tagging tasks.
\newblock \emph{arXiv preprint arXiv:1801.07772}.

\bibitem[{Benton et~al.(2017)Benton, Mitchell, and
  Hovy}]{benton-etal-2017-multitask}
Adrian Benton, Margaret Mitchell, and Dirk Hovy. 2017.
\newblock \href {https://www.aclweb.org/anthology/E17-1015} {Multitask learning
  for mental health conditions with limited social media data}.
\newblock In \emph{Proceedings of the 15th Conference of the {E}uropean Chapter
  of the Association for Computational Linguistics: Volume 1, Long Papers},
  pages 152--162.

\bibitem[{Bitew et~al.(2019)Bitew, Bekoulis, Deleu, Sterckx, Zaporojets,
  Demeester, and Develder}]{bitew2019predicting}
Semere~Kiros Bitew, Ioannis Bekoulis, Johannes Deleu, Lucas Sterckx, Klim
  Zaporojets, Thomas Demeester, and Chris Develder. 2019.
\newblock \href {https://www.aclweb.org/anthology/W19-3019/} {Predicting
  suicide risk from online postings in reddit: the ugent-idlab submission to
  the clpysch 2019 shared task a}.
\newblock In \emph{CLPsych2019, the 6th Annual Workshop on Computational
  Linguistics and Clinical Psychology at NAACL-HLT 2019}, pages 158--161.
  Association for Computational Linguistics (ACL).

\bibitem[{Bridges et~al.(2010)Bridges, Harnish et~al.}]{bridges2010role}
K~Robert Bridges, Richard~J Harnish, et~al. 2010.
\newblock \href
  {https://www.researchgate.net/publication/265873783_Role_of_irrational_beliefs_in_depression_and_anxiety_A_review}
  {Role of irrational beliefs in depression and anxiety: a review}.
\newblock \emph{Health}, 2(08):862.

\bibitem[{Carr(2020)}]{carr2020ai}
Sarah Carr. 2020.
\newblock \href
  {https://www.tandfonline.com/doi/full/10.1080/09638237.2020.1714011} {‘ai
  gone mental’: engagement and ethics in data-driven technology for mental
  health}.

\bibitem[{Caruana et~al.(2015)Caruana, Lou, Gehrke, Koch, Sturm, and
  Elhadad}]{caruana2015intelligible}
Rich Caruana, Yin Lou, Johannes Gehrke, Paul Koch, Marc Sturm, and Noemie
  Elhadad. 2015.
\newblock \href
  {https://www.microsoft.com/en-us/research/wp-content/uploads/2017/06/KDD2015FinalDraftIntelligibleModels4HealthCare_igt143e-caruanaA.pdf}
  {Intelligible models for healthcare: Predicting pneumonia risk and hospital
  30-day readmission}.
\newblock In \emph{Proceedings of the 21th ACM SIGKDD International Conference
  on Knowledge Discovery and Data Mining}, pages 1721--1730. ACM.

\bibitem[{Cer et~al.(2018)Cer, Yang, Kong, Hua, Limtiaco, John, Constant,
  Guajardo-C{\'e}spedes, Yuan, Tar et~al.}]{cer2018universal}
Daniel Cer, Yinfei Yang, Sheng-yi Kong, Nan Hua, Nicole Limtiaco, Rhomni~St
  John, Noah Constant, Mario Guajardo-C{\'e}spedes, Steve Yuan, Chris Tar,
  et~al. 2018.
\newblock \href {https://arxiv.org/abs/1803.11175} {Universal sentence
  encoder}.
\newblock \emph{arXiv preprint arXiv:1803.11175}.

\bibitem[{Chen et~al.(2019)Chen, Aldayel, Bogoychev, and
  Gong}]{chen2019similar}
Lushi Chen, Abeer Aldayel, Nikolay Bogoychev, and Tao Gong. 2019.
\newblock \href {https://www.aclweb.org/anthology/W19-3018/} {Similar minds
  post alike: Assessment of suicide risk using a hybrid model}.
\newblock In \emph{Proceedings of the Sixth Workshop on Computational
  Linguistics and Clinical Psychology}, pages 152--157.

\bibitem[{Chi et~al.(2020)Chi, Hewitt, and Manning}]{chi-etal-2020-finding}
Ethan~A. Chi, John Hewitt, and Christopher~D. Manning. 2020.
\newblock \href {https://doi.org/10.18653/v1/2020.acl-main.493} {Finding
  universal grammatical relations in multilingual {BERT}}.
\newblock In \emph{Proceedings of the 58th Annual Meeting of the Association
  for Computational Linguistics}, pages 5564--5577.

\bibitem[{Cohan et~al.(2018)Cohan, Desmet, Yates, Soldaini, MacAvaney, and
  Goharian}]{cohan2018smhd}
Arman Cohan, Bart Desmet, Andrew Yates, Luca Soldaini, Sean MacAvaney, and
  Nazli Goharian. 2018.
\newblock \href {https://www.aclweb.org/anthology/C18-1126/} {Smhd: a
  large-scale resource for exploring online language usage for multiple mental
  health conditions}.
\newblock In \emph{Proceedings of the 27th International Conference on
  Computational Linguistics}, pages 1485--1497.

\bibitem[{Coppersmith et~al.(2014)Coppersmith, Dredze, and
  Harman}]{coppersmith2014quantifying}
Glen Coppersmith, Mark Dredze, and Craig Harman. 2014.
\newblock \href {https://www.aclweb.org/anthology/W14-3207/} {Quantifying
  mental health signals in twitter}.
\newblock In \emph{Proceedings of the workshop on computational linguistics and
  clinical psychology: From linguistic signal to clinical reality}, pages
  51--60.

\bibitem[{Coppersmith et~al.(2015)Coppersmith, Dredze, Harman, Hollingshead,
  and Mitchell}]{coppersmith2015clpsych}
Glen Coppersmith, Mark Dredze, Craig Harman, Kristy Hollingshead, and Margaret
  Mitchell. 2015.
\newblock \href {https://www.aclweb.org/anthology/W15-1204.pdf} {Clpsych 2015
  shared task: Depression and ptsd on twitter}.
\newblock In \emph{Proceedings of the 2nd Workshop on Computational Linguistics
  and Clinical Psychology: From Linguistic Signal to Clinical Reality}, pages
  31--39.

\bibitem[{Czeisler et~al.(2020)Czeisler, Lane, Petrosky, Wiley, Christensen,
  Njai, Weaver, Robbins, Facer-Childs, Barger et~al.}]{czeisler2020mental}
Mark~{\'E} Czeisler, Rashon~I Lane, Emiko Petrosky, Joshua~F Wiley, Aleta
  Christensen, Rashid Njai, Matthew~D Weaver, Rebecca Robbins, Elise~R
  Facer-Childs, Laura~K Barger, et~al. 2020.
\newblock \href {https://www.ncbi.nlm.nih.gov/pmc/articles/PMC7440121/} {Mental
  health, substance use, and suicidal ideation during the covid-19
  pandemic—united states, june 24--30, 2020}.
\newblock \emph{Morbidity and Mortality Weekly Report}, 69(32):1049.

\bibitem[{Deng et~al.(2020)Deng, Ji, Rainey, Zhang, and
  Lu}]{deng2020integrating}
Changyu Deng, Xunbi Ji, Colton Rainey, Jianyu Zhang, and Wei Lu. 2020.
\newblock \href
  {https://www.sciencedirect.com/science/article/pii/S2589004220308488#sec2.2}
  {Integrating machine learning with human knowledge}.
\newblock \emph{Iscience}, page 101656.

\bibitem[{Devlin et~al.(2019)Devlin, Chang, Lee, and
  Toutanova}]{devlin-etal-2019-bert}
Jacob Devlin, Ming-Wei Chang, Kenton Lee, and Kristina Toutanova. 2019.
\newblock \href {https://doi.org/10.18653/v1/N19-1423} {{BERT}: Pre-training of
  deep bidirectional transformers for language understanding}.
\newblock In \emph{Proceedings of the 2019 Conference of the North {A}merican
  Chapter of the Association for Computational Linguistics: Human Language
  Technologies, Volume 1 (Long and Short Papers)}, pages 4171--4186.

\bibitem[{Ettman et~al.(2020)Ettman, Abdalla, Cohen, Sampson, Vivier, and
  Galea}]{ettman2020prevalence}
Catherine~K Ettman, Salma~M Abdalla, Gregory~H Cohen, Laura Sampson, Patrick~M
  Vivier, and Sandro Galea. 2020.
\newblock \href
  {https://jamanetwork.com/journals/jamanetworkopen/fullarticle/2770146/}
  {Prevalence of depression symptoms in us adults before and during the
  covid-19 pandemic}.
\newblock \emph{JAMA network open}, 3(9):e2019686--e2019686.

\bibitem[{Guntuku et~al.(2017)Guntuku, Yaden, Kern, Ungar, and
  Eichstaedt}]{guntuku2017detecting}
Sharath~Chandra Guntuku, David~B Yaden, Margaret~L Kern, Lyle~H Ungar, and
  Johannes~C Eichstaedt. 2017.
\newblock \href
  {https://www.semanticscholar.org/paper/Detecting-depression-and-mental-illness-on-social-Guntuku-Yaden/c6bbfcfc16e3dcb33b0463a95df44d543838c76d}
  {Detecting depression and mental illness on social media: an integrative
  review}.
\newblock \emph{Current Opinion in Behavioral Sciences}, 18:43--49.

\bibitem[{Hancock et~al.(2018)Hancock, Bringmann, Varma, Liang, Wang, and
  R{\'e}}]{hancock2018training}
Braden Hancock, Martin Bringmann, Paroma Varma, Percy Liang, Stephanie Wang,
  and Christopher R{\'e}. 2018.
\newblock \href {https://www.aclweb.org/anthology/P18-1175/} {Training
  classifiers with natural language explanations}.
\newblock In \emph{Proceedings of the conference. Association for Computational
  Linguistics. Meeting}, volume 2018, page 1884. NIH Public Access.

\bibitem[{Harrigian et~al.(2020)Harrigian, Aguirre, and
  Dredze}]{harrigian2020models}
Keith Harrigian, Carlos Aguirre, and Mark Dredze. 2020.
\newblock \href {https://www.aclweb.org/anthology/2020.findings-emnlp.337/} {Do
  models of mental health based on social media data generalize?}
\newblock In \emph{Proceedings of the 2020 Conference on Empirical Methods in
  Natural Language Processing: Findings (EMNLP)}, pages 3774--3788.

\bibitem[{Heaton et~al.(2016)Heaton, Polson, and Witte}]{heaton2016deep}
JB~Heaton, Nicholas~G Polson, and Jan~Hendrik Witte. 2016.
\newblock \href {https://arxiv.org/abs/1602.06561} {Deep learning in finance}.
\newblock \emph{arXiv preprint arXiv:1602.06561}.

\bibitem[{Hevia et~al.(2019)Hevia, Men{\'e}ndez, and
  Gayo-Avello}]{hevia2019analyzing}
Alejandro~Gonz{\'a}lez Hevia, Rebeca~Cerezo Men{\'e}ndez, and Daniel
  Gayo-Avello. 2019.
\newblock \href {https://www.aclweb.org/anthology/W19-3017/} {Analyzing the use
  of existing systems for the clpsych 2019 shared task}.
\newblock In \emph{Proceedings of the Sixth Workshop on Computational
  Linguistics and Clinical Psychology}, pages 148--151.

\bibitem[{Hewitt and Manning(2019)}]{hewitt2019structural}
John Hewitt and Christopher~D Manning. 2019.
\newblock A structural probe for finding syntax in word representations.
\newblock In \emph{Proceedings of the 2019 Conference of the North American
  Chapter of the Association for Computational Linguistics: Human Language
  Technologies, Volume 1 (Long and Short Papers)}, pages 4129--4138.

\bibitem[{Iserman et~al.(2019)Iserman, Nalabandian, and
  Ireland}]{iserman2019dictionaries}
Micah Iserman, Taleen Nalabandian, and Molly Ireland. 2019.
\newblock \href {https://www.aclweb.org/anthology/W19-3025/} {Dictionaries and
  decision trees for the 2019 clpsych shared task}.
\newblock In \emph{Proceedings of the Sixth Workshop on Computational
  Linguistics and Clinical Psychology}, pages 188--194.

\bibitem[{Kang et~al.(2018)Kang, Zhang, Kummerfeld, Tang, and
  Mars}]{kang2018data}
Yiping Kang, Yunqi Zhang, Jonathan~K Kummerfeld, Lingjia Tang, and Jason Mars.
  2018.
\newblock \href {https://www.aclweb.org/anthology/N18-3005/} {Data collection
  for dialogue system: A startup perspective}.
\newblock In \emph{Proceedings of the 2018 Conference of the North American
  Chapter of the Association for Computational Linguistics: Human Language
  Technologies, Volume 3 (Industry Papers)}, pages 33--40.

\bibitem[{Kaplan et~al.(2017)Kaplan, Morrison, Goldin, Olino, Heimberg, and
  Gross}]{kaplan2017cognitive}
Simona~C Kaplan, Amanda~S Morrison, Philippe~R Goldin, Thomas~M Olino,
  Richard~G Heimberg, and James~J Gross. 2017.
\newblock \href {https://pubmed.ncbi.nlm.nih.gov/28966414/} {The cognitive
  distortions questionnaire (cd-quest): Validation in a sample of adults with
  social anxiety disorder}.
\newblock \emph{Cognitive therapy and research}, 41(4):576--587.

\bibitem[{Kroenke et~al.(2001)Kroenke, Spitzer, and Williams}]{kroenke2001phq}
Kurt Kroenke, Robert~L Spitzer, and Janet~BW Williams. 2001.
\newblock \href {https://pubmed.ncbi.nlm.nih.gov/11556941/} {The phq-9:
  validity of a brief depression severity measure}.
\newblock \emph{Journal of general internal medicine}, 16(9):606--613.

\bibitem[{Larson et~al.(2019)Larson, Mahendran, Lee, Kummerfeld, Johann, and
  Mars}]{larson2019outlier}
Stefan Larson, Anish Mahendran, Andrew Lee, Jonathan~K Kummerfeld, Parker Hill
  Michael A~Laurenzano Johann, and Hauswald Lingjia Tang~Jason Mars. 2019.
\newblock \href {https://www.aclweb.org/anthology/N19-1051.pdf} {Outlier
  detection for improved data quality and diversity in dialog systems}.
\newblock In \emph{Proceedings of NAACL-HLT}, pages 517--527.

\bibitem[{Larson et~al.(2020)Larson, Zheng, Mahendran, Tekriwal, Cheung,
  Guldan, Leach, and Kummerfeld}]{larson-etal-2020-iterative}
Stefan Larson, Anthony Zheng, Anish Mahendran, Rishi Tekriwal, Adrian Cheung,
  Eric Guldan, Kevin Leach, and Jonathan~K. Kummerfeld. 2020.
\newblock \href {https://doi.org/10.18653/v1/2020.emnlp-main.650} {Iterative
  feature mining for constraint-based data collection to increase data
  diversity and model robustness}.
\newblock In \emph{Proceedings of the 2020 Conference on Empirical Methods in
  Natural Language Processing (EMNLP)}, pages 8097--8106.

\bibitem[{Levy and Deykin(1989)}]{levy1989suicidality}
Janice~C Levy and Eva~Y Deykin. 1989.
\newblock \href {https://pubmed.ncbi.nlm.nih.gov/2817119/} {Suicidality,
  depression, and substance abuse in adolescence.}
\newblock \emph{The American journal of psychiatry}.

\bibitem[{Liu et~al.(2020)Liu, Zhou, Zhao, Wang, Ju, Deng, and Wang}]{liu2020k}
Weijie Liu, Peng Zhou, Zhe Zhao, Zhiruo Wang, Qi~Ju, Haotang Deng, and Ping
  Wang. 2020.
\newblock \href {https://ojs.aaai.org/index.php/AAAI/article/view/5681/5537}
  {K-bert: Enabling language representation with knowledge graph}.
\newblock In \emph{Proceedings of the AAAI Conference on Artificial
  Intelligence}, volume~34, pages 2901--2908.

\bibitem[{Lou et~al.(2012)Lou, Caruana, and Gehrke}]{lou2012intelligible}
Yin Lou, Rich Caruana, and Johannes Gehrke. 2012.
\newblock \href {https://www.cs.cornell.edu/~yinlou/papers/lou-kdd12.pdf}
  {Intelligible models for classification and regression}.
\newblock In \emph{Proceedings of the 18th ACM SIGKDD international conference
  on Knowledge discovery and data mining}, pages 150--158.

\bibitem[{Lou et~al.(2013)Lou, Caruana, Gehrke, and Hooker}]{lou2013accurate}
Yin Lou, Rich Caruana, Johannes Gehrke, and Giles Hooker. 2013.
\newblock \href {https://www.cs.cornell.edu/~yinlou/papers/lou-kdd13.pdf}
  {Accurate intelligible models with pairwise interactions}.
\newblock In \emph{Proceedings of the 19th ACM SIGKDD international conference
  on Knowledge discovery and data mining}, pages 623--631.

\bibitem[{Lundberg and Lee(2017)}]{lundberg2017unified}
Scott Lundberg and Su-In Lee. 2017.
\newblock \href {https://arxiv.org/abs/1705.07874} {A unified approach to
  interpreting model predictions}.
\newblock \emph{arXiv preprint arXiv:1705.07874}.

\bibitem[{Matero et~al.(2019)Matero, Idnani, Son, Giorgi, Vu, Zamani,
  Limbachiya, Guntuku, and Schwartz}]{matero2019suicide}
Matthew Matero, Akash Idnani, Youngseo Son, Salvatore Giorgi, Huy Vu, Mohammad
  Zamani, Parth Limbachiya, Sharath~Chandra Guntuku, and H~Andrew Schwartz.
  2019.
\newblock \href {https://www.aclweb.org/anthology/W19-3005/} {Suicide risk
  assessment with multi-level dual-context language and bert}.
\newblock In \emph{Proceedings of the Sixth Workshop on Computational
  Linguistics and Clinical Psychology}, pages 39--44.

\bibitem[{Mohammadi et~al.(2019)Mohammadi, Amini, and
  Kosseim}]{mohammadi2019clac}
Elham Mohammadi, Hessam Amini, and Leila Kosseim. 2019.
\newblock \href {https://www.aclweb.org/anthology/W19-3004/} {Clac at clpsych
  2019: Fusion of neural features and predicted class probabilities for suicide
  risk assessment based on online posts}.
\newblock In \emph{Proceedings of the Sixth Workshop on Computational
  Linguistics and Clinical Psychology}, pages 34--38.

\bibitem[{Morales et~al.(2019)Morales, Dey, Theisen, Belitz, and
  Chernova}]{morales2019investigation}
Michelle Morales, Prajjalita Dey, Thomas Theisen, Daniel Belitz, and Natalia
  Chernova. 2019.
\newblock \href {https://www.aclweb.org/anthology/W19-3023/} {An investigation
  of deep learning systems for suicide risk assessment}.
\newblock In \emph{Proceedings of the sixth workshop on computational
  linguistics and clinical psychology}, pages 177--181.

\bibitem[{Nadareishvili et~al.(2016)Nadareishvili, Mitra, McLarty, and
  Amundsen}]{nadareishvili2016microservice}
Irakli Nadareishvili, Ronnie Mitra, Matt McLarty, and Mike Amundsen. 2016.
\newblock \emph{Microservice architecture: aligning principles, practices, and
  culture}.
\newblock " O'Reilly Media, Inc.".

\bibitem[{Nori et~al.(2019)Nori, Jenkins, Koch, and
  Caruana}]{nori2019interpretml}
Harsha Nori, Samuel Jenkins, Paul Koch, and Rich Caruana. 2019.
\newblock \href {https://arxiv.org/abs/1909.09223} {Interpretml: A unified
  framework for machine learning interpretability}.
\newblock \emph{arXiv preprint arXiv:1909.09223}.

\bibitem[{Orabi et~al.(2018)Orabi, Buddhitha, Orabi, and
  Inkpen}]{orabi2018deep}
Ahmed~Husseini Orabi, Prasadith Buddhitha, Mahmoud~Husseini Orabi, and Diana
  Inkpen. 2018.
\newblock \href {https://www.aclweb.org/anthology/W18-0609/} {Deep learning for
  depression detection of twitter users}.
\newblock In \emph{Proceedings of the Fifth Workshop on Computational
  Linguistics and Clinical Psychology: From Keyboard to Clinic}, pages 88--97.

\bibitem[{Pennebaker et~al.(2001)Pennebaker, Francis, and
  Booth}]{pennebaker2001linguistic}
James~W Pennebaker, Martha~E Francis, and Roger~J Booth. 2001.
\newblock \href
  {https://www.cs.cmu.edu/~ylataus/files/TausczikPennebaker2010.pdf}
  {Linguistic inquiry and word count: Liwc 2001}.
\newblock \emph{Mahway: Lawrence Erlbaum Associates}, 71(2001):2001.

\bibitem[{Preotiuc-Pietro et~al.(2015)Preotiuc-Pietro, Sap, Schwartz, and
  Ungar}]{preotiuc2015mental}
Daniel Preotiuc-Pietro, Maarten Sap, H~Andrew Schwartz, and Lyle~H Ungar. 2015.
\newblock \href {https://www.aclweb.org/anthology/W15-1205.pdf} {Mental illness
  detection at the world well-being project for the clpsych 2015 shared task.}
\newblock In \emph{CLPsych@ HLT-NAACL}, pages 40--45.

\bibitem[{Ratner et~al.(2016)Ratner, De~Sa, Wu, Selsam, and
  R{\'e}}]{ratner2016data}
Alexander Ratner, Christopher De~Sa, Sen Wu, Daniel Selsam, and Christopher
  R{\'e}. 2016.
\newblock \href
  {https://papers.nips.cc/paper/2016/file/6709e8d64a5f47269ed5cea9f625f7ab-Paper.pdf}
  {Data programming: Creating large training sets, quickly}.
\newblock \emph{Advances in neural information processing systems}, 29:3567.

\bibitem[{Ratner et~al.(2017)Ratner, Bach, Ehrenberg, and
  R{\'e}}]{ratner2017snorkel}
Alexander~J Ratner, Stephen~H Bach, Henry~R Ehrenberg, and Chris R{\'e}. 2017.
\newblock \href {https://dl.acm.org/doi/10.1145/3035918.3056442} {Snorkel: Fast
  training set generation for information extraction}.
\newblock In \emph{Proceedings of the 2017 ACM international conference on
  management of data}, pages 1683--1686.

\bibitem[{Reimers and Gurevych(2019)}]{reimers-gurevych-2019-sentence}
Nils Reimers and Iryna Gurevych. 2019.
\newblock \href {https://doi.org/10.18653/v1/D19-1410} {Sentence-{BERT}:
  Sentence embeddings using {S}iamese {BERT}-networks}.
\newblock In \emph{Proceedings of the 2019 Conference on Empirical Methods in
  Natural Language Processing and the 9th International Joint Conference on
  Natural Language Processing (EMNLP-IJCNLP)}, pages 3982--3992.

\bibitem[{Resnik et~al.(2015)Resnik, Armstrong, Claudino, and
  Nguyen}]{resnik2015university}
Philip Resnik, William Armstrong, Leonardo Claudino, and Thang Nguyen. 2015.
\newblock \href {https://www.aclweb.org/anthology/W15-1207/} {The university of
  maryland clpsych 2015 shared task system}.
\newblock In \emph{Proceedings of the 2nd workshop on computational linguistics
  and clinical psychology: from linguistic signal to clinical reality}, pages
  54--60.

\bibitem[{Ribeiro et~al.(2016)Ribeiro, Singh, and Guestrin}]{ribeiro2016should}
Marco~Tulio Ribeiro, Sameer Singh, and Carlos Guestrin. 2016.
\newblock \href {https://dl.acm.org/doi/abs/10.1145/2939672.2939778} {" why
  should i trust you?" explaining the predictions of any classifier}.
\newblock In \emph{Proceedings of the 22nd ACM SIGKDD international conference
  on knowledge discovery and data mining}, pages 1135--1144.

\bibitem[{R{\'\i}ssola et~al.(2019)R{\'\i}ssola, Ram{\'\i}{\i}rez-Cifuentes,
  Freire, and Crestani}]{rissola2019suicide}
Esteban~A R{\'\i}ssola, Diana Ram{\'\i}{\i}rez-Cifuentes, Ana Freire, and Fabio
  Crestani. 2019.
\newblock \href {https://www.aclweb.org/anthology/W19-3021/} {Suicide risk
  assessment on social media: Usi-upf at the clpsych 2019 shared task}.
\newblock In \emph{Proceedings of the Sixth Workshop on Computational
  Linguistics and Clinical Psychology: 2019 Jun 6; Minneapolis, Minnesota, USA.
  Stroudsburg: ACL; 2019. p. 167--71.} ACL (Association for Computational
  Linguistics).

\bibitem[{Ruggiero et~al.(2003)Ruggiero, Del~Ben, Scotti, and
  Rabalais}]{ruggiero2003psychometric}
Kenneth~J Ruggiero, Kevin Del~Ben, Joseph~R Scotti, and Aline~E Rabalais. 2003.
\newblock \href {https://link.springer.com/article/10.1023/A:1025714729117}
  {Psychometric properties of the ptsd checklist—civilian version}.
\newblock \emph{Journal of traumatic stress}, 16(5):495--502.

\bibitem[{Ruiz et~al.(2019)Ruiz, Shi, Quan, Ryan, Biernesser, Brent, and
  Tsui}]{ruiz2019clpsych2019}
Victor Ruiz, Lingyun Shi, Wei Quan, Neal Ryan, Candice Biernesser, David Brent,
  and Rich Tsui. 2019.
\newblock \href {https://www.aclweb.org/anthology/W19-3020/} {Clpsych2019
  shared task: Predicting suicide risk level from reddit posts on multiple
  forums}.
\newblock In \emph{Proceedings of the Sixth Workshop on Computational
  Linguistics and Clinical Psychology}, pages 162--166.

\bibitem[{Sastre-Buades et~al.(2021)Sastre-Buades, Ochoa, Lorente-Rovira,
  Barajas, Grasa, L{\'o}pez-Carrilero, Luengo, Ruiz-Delgado, Cid,
  Gonz{\'a}lez-Higueras et~al.}]{sastre2021jumping}
Aina Sastre-Buades, Susana Ochoa, Esther Lorente-Rovira, Ana Barajas, Eva
  Grasa, Raquel L{\'o}pez-Carrilero, Ana Luengo, Isabel Ruiz-Delgado, Jordi
  Cid, Ferm{\'\i}n Gonz{\'a}lez-Higueras, et~al. 2021.
\newblock \href
  {https://www.sciencedirect.com/science/article/pii/S0022395621001680}
  {Jumping to conclusions and suicidal behavior in depression and psychosis}.
\newblock \emph{Journal of psychiatric research}.

\bibitem[{Shing et~al.(2018)Shing, Nair, Zirikly, Friedenberg, {Daum{\'e} III},
  and Resnik}]{shing2018expert}
Han-Chin Shing, Suraj Nair, Ayah Zirikly, Meir Friedenberg, Hal {Daum{\'e}
  III}, and Philip Resnik. 2018.
\newblock \href {https://www.aclweb.org/anthology/W18-0603/} {Expert,
  crowdsourced, and machine assessment of suicide risk via online postings}.
\newblock In \emph{Proceedings of the Fifth Workshop on Computational
  Linguistics and Clinical Psychology: From Keyboard to Clinic}, pages 25--36.

\bibitem[{Spitzer et~al.(2006)Spitzer, Kroenke, Williams, and
  L{\"o}we}]{spitzer2006brief}
Robert~L Spitzer, Kurt Kroenke, Janet~BW Williams, and Bernd L{\"o}we. 2006.
\newblock \href {https://pubmed.ncbi.nlm.nih.gov/16717171/} {A brief measure
  for assessing generalized anxiety disorder: the gad-7}.
\newblock \emph{Archives of internal medicine}, 166(10):1092--1097.

\bibitem[{Stasaski et~al.(2020)Stasaski, Yang, and
  Hearst}]{stasaski-etal-2020-diverse}
Katherine Stasaski, Grace~Hui Yang, and Marti~A. Hearst. 2020.
\newblock \href {https://doi.org/10.18653/v1/2020.acl-main.446} {More diverse
  dialogue datasets via diversity-informed data collection}.
\newblock In \emph{Proceedings of the 58th Annual Meeting of the Association
  for Computational Linguistics}, pages 4958--4968.

\bibitem[{Swearer et~al.(2001)Swearer, Song, Cary, Eagle, and
  Mickelson}]{swearer2001psychosocial}
Susan~M Swearer, Samuel~Y Song, Paulette~Tam Cary, John~W Eagle, and William~T
  Mickelson. 2001.
\newblock \href {https://www.tandfonline.com/doi/abs/10.1300/J135v02n02_07}
  {Psychosocial correlates in bullying and victimization: The relationship
  between depression, anxiety, and bully/victim status}.
\newblock \emph{Journal of Emotional Abuse}, 2(2-3):95--121.

\bibitem[{Tenney et~al.(2018)Tenney, Xia, Chen, Wang, Poliak, McCoy, Kim,
  Van~Durme, Bowman, Das et~al.}]{tenney2018you}
Ian Tenney, Patrick Xia, Berlin Chen, Alex Wang, Adam Poliak, R~Thomas McCoy,
  Najoung Kim, Benjamin Van~Durme, Samuel~R Bowman, Dipanjan Das, et~al. 2018.
\newblock What do you learn from context? probing for sentence structure in
  contextualized word representations.
\newblock In \emph{International Conference on Learning Representations}.

\bibitem[{Vig et~al.(2020)Vig, Gehrmann, Belinkov, Qian, Nevo, Singer, and
  Shieber}]{vig2020investigating}
Jesse Vig, Sebastian Gehrmann, Yonatan Belinkov, Sharon Qian, Daniel Nevo,
  Yaron Singer, and Stuart~M Shieber. 2020.
\newblock Investigating gender bias in language models using causal mediation
  analysis.
\newblock In \emph{NeurIPS}.

\bibitem[{Winata et~al.(2018)Winata, Kampman, and Fung}]{winata2018attention}
Genta~Indra Winata, Onno~Pepijn Kampman, and Pascale Fung. 2018.
\newblock \href {https://ieeexplore.ieee.org/abstract/document/8461990}
  {Attention-based lstm for psychological stress detection from spoken language
  using distant supervision}.
\newblock In \emph{2018 IEEE International Conference on Acoustics, Speech and
  Signal Processing (ICASSP)}, pages 6204--6208. IEEE.

\bibitem[{Wu et~al.(2021)Wu, Peng, and Smith}]{wu2021infusing}
Zhaofeng Wu, Hao Peng, and Noah Smith. 2021.
\newblock Infusing finetuning with semantic dependencies.
\newblock \emph{Transactions of the Association for Computational Linguistics},
  9:226--242.

\bibitem[{Xie et~al.(2020)Xie, Niu, Liu, Chen, and Tang}]{xie2020survey}
Xiaozheng Xie, Jianwei Niu, Xuefeng Liu, Zhengsu Chen, and Shaojie Tang. 2020.
\newblock A survey on domain knowledge powered deep learning for medical image
  analysis.
\newblock \emph{arXiv preprint arXiv:2004.12150}.

\bibitem[{Yang et~al.(2019)Yang, Al-Bahrani, Reid, Papanikolaou, Kalidindi,
  Liao, Choudhary, and Agrawal}]{yang2019deep}
Zijiang Yang, Reda Al-Bahrani, Andrew~CE Reid, Stefanos Papanikolaou, Surya~R
  Kalidindi, Wei-keng Liao, Alok Choudhary, and Ankit Agrawal. 2019.
\newblock \href {https://ieeexplore.ieee.org/abstract/document/8852162} {Deep
  learning based domain knowledge integration for small datasets: Illustrative
  applications in materials informatics}.
\newblock In \emph{2019 International Joint Conference on Neural Networks
  (IJCNN)}, pages 1--8. IEEE.

\bibitem[{Yates et~al.(2017)Yates, Cohan, and
  Goharian}]{yates-etal-2017-depression}
Andrew Yates, Arman Cohan, and Nazli Goharian. 2017.
\newblock \href {https://doi.org/10.18653/v1/D17-1322} {Depression and
  self-harm risk assessment in online forums}.
\newblock In \emph{Proceedings of the 2017 Conference on Empirical Methods in
  Natural Language Processing (EMNLP)}, pages 2968--2978.

\bibitem[{Zahn et~al.(2015)Zahn, Lythe, Gethin, Green, Deakin, Young, and
  Moll}]{zahn2015role}
Roland Zahn, Karen~E Lythe, Jennifer~A Gethin, Sophie Green, John F~William
  Deakin, Allan~H Young, and Jorge Moll. 2015.
\newblock \href {https://www.ncbi.nlm.nih.gov/pmc/articles/PMC4573463/} {The
  role of self-blame and worthlessness in the psychopathology of major
  depressive disorder}.
\newblock \emph{Journal of affective disorders}, 186:337--341.

\bibitem[{Zhang et~al.(2019)Zhang, Han, Liu, Jiang, Sun, and
  Liu}]{zhang2019ernie}
Zhengyan Zhang, Xu~Han, Zhiyuan Liu, Xin Jiang, Maosong Sun, and Qun Liu. 2019.
\newblock \href {https://www.aclweb.org/anthology/P19-1139.pdf} {Ernie:
  Enhanced language representation with informative entities}.
\newblock In \emph{Proceedings of the 57th Annual Meeting of the Association
  for Computational Linguistics}, pages 1441--1451.

\bibitem[{Zirikly et~al.(2019)Zirikly, Resnik, Uzuner, and
  Hollingshead}]{zirikly2019clpsych}
Ayah Zirikly, Philip Resnik, Ozlem Uzuner, and Kristy Hollingshead. 2019.
\newblock \href {https://www.aclweb.org/anthology/W19-3003/} {Clpsych 2019
  shared task: Predicting the degree of suicide risk in reddit posts}.
\newblock In \emph{Proceedings of the sixth workshop on computational
  linguistics and clinical psychology}, pages 24--33.

\end{thebibliography}


\begin{thebibliography}{2}
\expandafter\ifx\csname natexlab\endcsname\relax\def\natexlab#1{#1}\fi

\bibitem[{Burns and Beck(1999)}]{burns1999feeling}
David~D Burns and Aaron~T Beck. 1999.
\newblock Feeling good: The new mood therapy.

\bibitem[{Kroenke et~al.(2001)Kroenke, Spitzer, and Williams}]{kroenke2001phq}
Kurt Kroenke, Robert~L Spitzer, and Janet~BW Williams. 2001.
\newblock \href {https://pubmed.ncbi.nlm.nih.gov/11556941/} {The phq-9:
  validity of a brief depression severity measure}.
\newblock \emph{Journal of general internal medicine}, 16(9):606--613.

\end{thebibliography}
\bibliographystyle{acl_natbib}

\appendix

\end{document}


\appendix

\section{Feature Importance Scores}
\label{appendix:feature_importance_scores}

\begin{figure}
\centering\includegraphics[width=0.95\columnwidth]{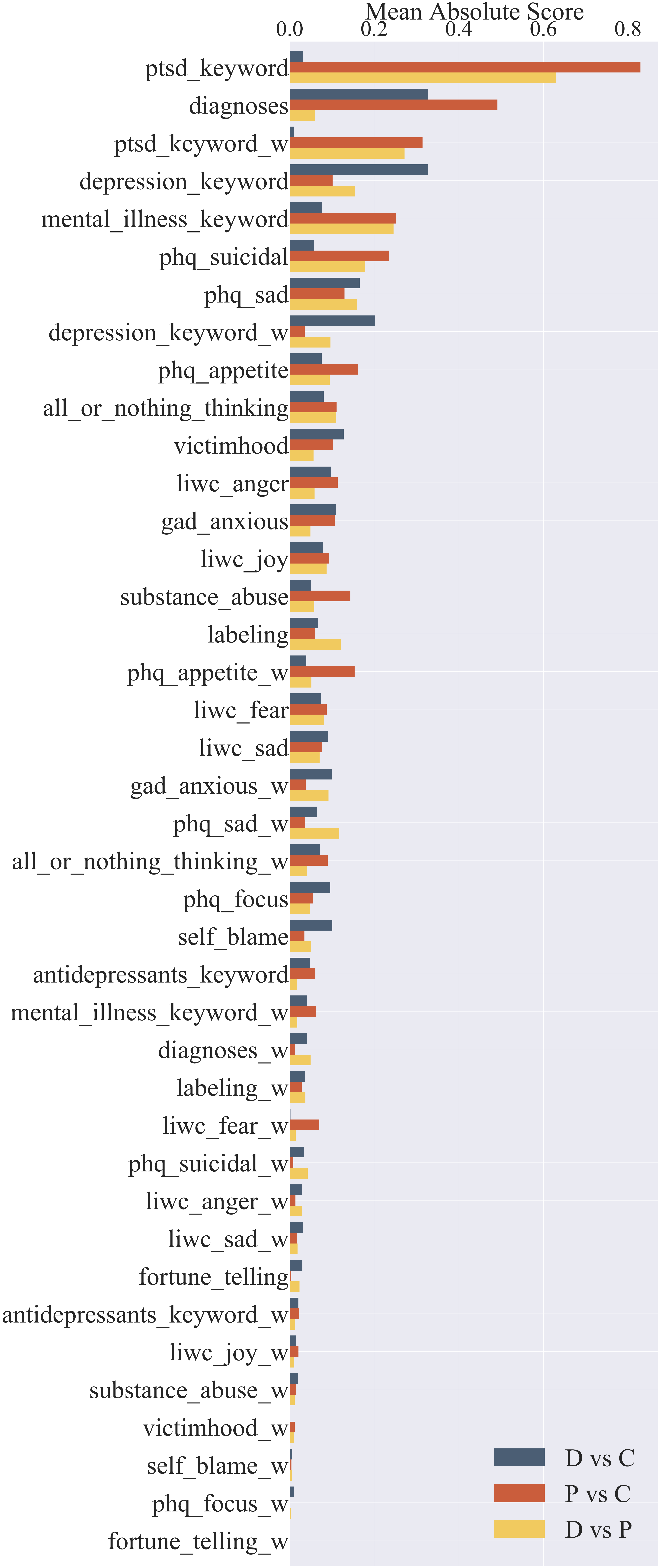}
\caption{\label{appx_fig_feature_importance}
Global feature importance scores of the EBM classifiers trained on depression vs condition, PTSD vs condition, and depression vs PTSD.
Features ending in "w" are features from the aggregator that looks for windows of hits.
}
\end{figure}

Figure~\ref{appx_fig_feature_importance} lists the global feature importance scores for the features used in classifying 1) depression versus control, 2) PTSD versus control, and 3) depression versus PTSD.

\begin{table*}
\resizebox{\textwidth}{!}{
    \centering
    \begin{tabular}{p{0.3\linewidth}|p{0.4\linewidth}|p{0.4\linewidth}}
        Name       & Description & Examples  \\
        \toprule
        All-or-Nothing Thinking & Seeing things in extreme, black-and-white categories. Thinking in absolutes such as "always", "never", or "every".          &  "I'm a \emph{total} failure." \newline "I \emph{never} do \emph{anything} right." \\
        
        Overgeneralization & Seeing a single negative event as a never-ending pattern of defeat. & "She said no -- I'm never going to get a date. I'll be lonely all my life." \newline "I didn't get the job. I'll never find a job." \\
        
        Labeling & Creating a completely negative self-image based on one's errors. Attaching a negative label to oneself. & "I'm an idiot!" \newline "I'm a loser." \\
        
        Fortune-Telling Error & Anticipating that things will turn out badly and feeling convinced that one's predictions are already-established facts. & "I'll make a fool of myself." \newline "I'll never get better." \\
        
        Disqualifying the Positive & Rejecting positive experiences by insisting they "don't count" for some reason or other. & (After a compliment) "They're just being nice." \newline "That was a fluke."

    \end{tabular}}
    \caption{\label{tab:cognitive_distortions}
    Definition and examples of common cognitive distortions according to \citet{burns1999feeling}}
\end{table*}

\section{Cognitive Distortions}
\label{appendix:cognitive_distortions}

Table~\ref{tab:cognitive_distortions} lists some common examples of cognitive distortions, along with their definitions and some examples.

\begin{table*}
    \centering
    \begin{tabular}{l}
        PHQ-9 Questionnaire \\
        \toprule
        1. Little interest or pleasure in doing things \\
        2. Feeling down, depressed, or hopeless \\
        3. Trouble falling or staying asleep, or sleeping too much \\
        4. Feeling tired or having little energy \\
        5. Poor appetite or overeating \\
        6. Feeling bad about yourself or that you are a failure or have let yourself or your family down \\
        7. Trouble concentrating on things, such as reading the newspaper or watching television \\
        8. Moving or speaking so slowly that other people could have noticed. \\
           Or the opposite being so figety or restless that you have been moving around a lot more than usual \\
        9. Thoughts that you would be better off dead, or of hurting yourself
\end{tabular}
\caption{\label{tab:phq_9}
PHQ-9 Questionnaire according to \citet{kroenke2001phq}}
\end{table*}

\section{PHQ-9 Questionnaire}
\label{appendix:phq9}

Table~\ref{tab:phq_9} lists the PHQ-9 Questionnaire.

\clearpage

\bibliography{custom}
\bibliographystyle{acl_natbib}